\documentclass[runningheads]{llncs}

\usepackage{eccv}

\usepackage{eccvabbrv}

\usepackage{graphicx}
\usepackage{booktabs}

\usepackage[accsupp]{axessibility}  

\usepackage[pagebackref,breaklinks,colorlinks,citecolor=eccvblue]{hyperref}

\usepackage{orcidlink}
\usepackage{xcolor}
\usepackage{colortbl}
\usepackage{animate}
\begin{document}

\title{Mobius: A High Efficient Spatial-Temporal Parallel Training Paradigm for Text-to-Video Generation Task} 

\titlerunning{Mobius}

\author{Yiran Yang\inst{1}\orcidlink{0000-0003-2688-6340 } \and
Jinchao Zhang\inst{2} \and
Ying Deng\inst{2} \and
Jie Zhou\inst{2}}

\authorrunning{Yang et al.}

\institute{Aerospace Information Research Institute, Chinese Academy of Sciences  \\
\email{yangyiran19@mails.ucas.ac.cn}
\and
Tencent, Wechat BG\\
\email{\{dayerzhang,cheryldeng, withtomzhou\}@tencent.com}}

\maketitle

\begin{abstract}
Inspired by the success of the text-to-image (T2I) generation task, many researchers are devoting themselves to the text-to-video (T2V) generation task. Most of the T2V frameworks usually inherit from the T2I model and add extra-temporal layers of training to generate dynamic videos, which can be viewed as a fine-tuning task. However, the traditional 3D-Unet is a serial mode and the temporal layers follow the spatial layers, which will result in high GPU memory and training time consumption according to its serial feature flow. We believe that this serial mode will bring more training costs with the large diffusion model and massive datasets, which are not environmentally friendly and not suitable for the development of the T2V. Therefore, we propose a highly efficient spatial-temporal parallel training paradigm for T2V tasks, named Mobius. In our 3D-Unet, the temporal layers and spatial layers are parallel, which optimizes the feature flow and backpropagation. The Mobius will save \textcolor{red}{\textbf{24\%}} GPU memory and \textcolor{red}{\textbf{12\%}} training time, which can greatly improve the T2V fine-tuning task and provide a novel insight for the AIGC community. We will release our \href{https://github.com/youngfly/Mobius}{codes} in the future.
  
\end{abstract}

\section{Introduction}

Artificial Intelligence Generated Content (AIGC) has emerged as a modern trend, utilizing advanced computational creation tools to revolutionize the traditional creation industry. Previous industry typically requires extensive resources, and an excellent artist involves artistic endeavors that necessitate years of training for realization. But now, intelligent creation not only empowers individuals to pursue their aspirations and achieve desired outcomes but also aids professionals in efficiently bringing their design concepts to life.

Diffusion models\cite{ho2020denoising,song2020denoising} have recently taken center stage in the AIGC domain, thanks to their superior generative capabilities. These models, which simulate the diffusion or spreading of particles, are gradually replacing methods based on Generative Adversarial Networks (GANs)\cite{goodfellow2020generative,brock2018large} and autoregressive Transformers\cite{chen2020generative,child2019generating,van2016conditional,van2016pixel,razavi2019generating,ramesh2021zero}. They have proven effective in generating complex, high-dimensional data such as images or video. In image generation\cite{rombach2022high}, diffusion models have exhibited impressive capabilities, generating realistic images by learning from existing image datasets and capturing intricate details and patterns. This makes them an invaluable tool for tasks like digital art creation, where they can produce unique, high-quality artworks. Furthermore, the application of diffusion models isn’t limited to static content creation. They have been used in dynamic fields like video generation and editing, where they can generate realistic image sequences, simulating motion and change over time.

\begin{figure}[!t]
	\centering
	\includegraphics[scale=0.35]{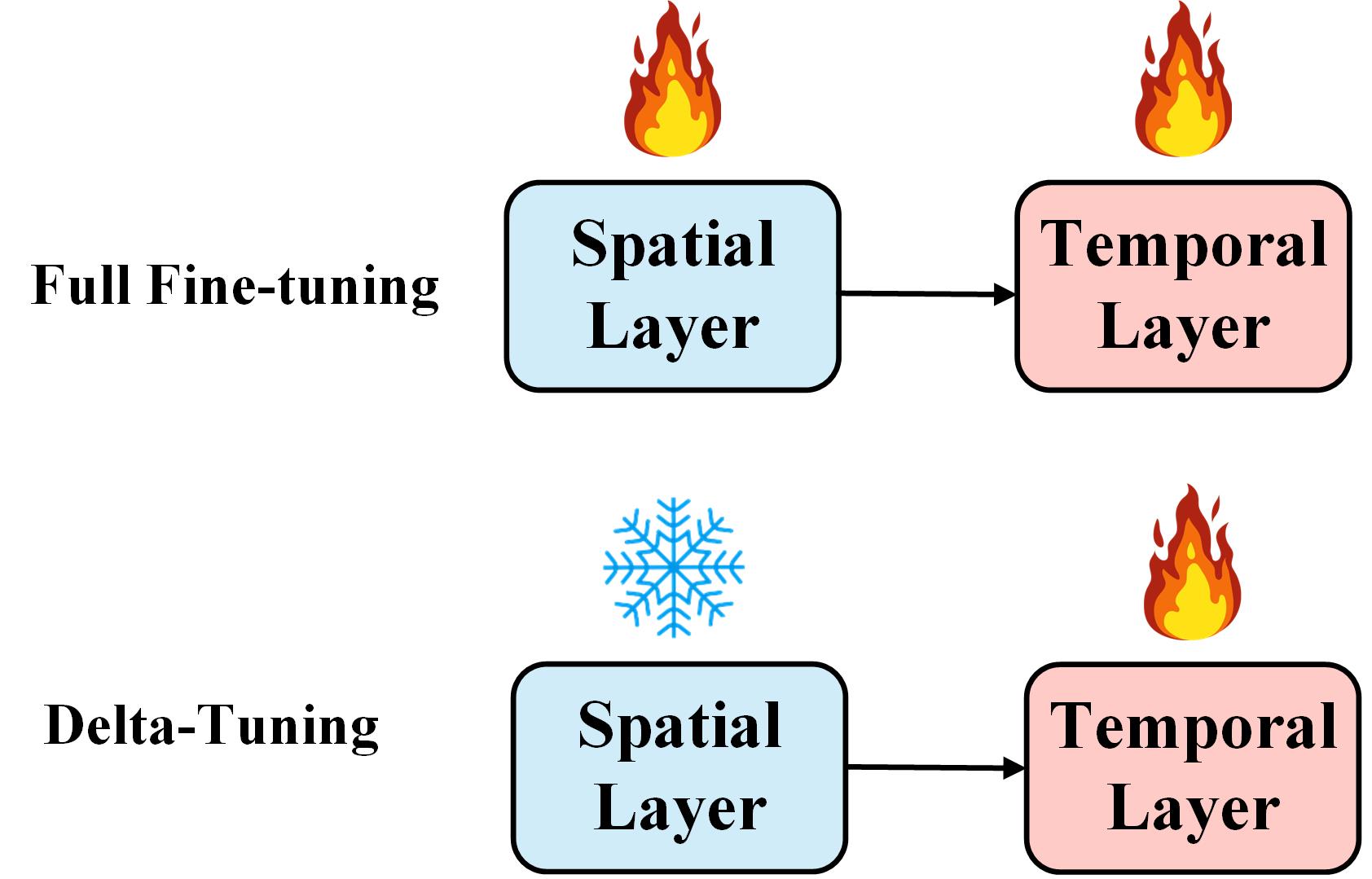}
	\caption{Existing tuning paradigm in T2V generation task. The top is the full fine-tuning which goes through all parameters. The bottom is Delta-Tuning which only update part parameters. \protect\includegraphics[height=0.3cm]{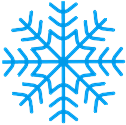} represents the frozen part, and \protect\includegraphics[height=0.3cm]{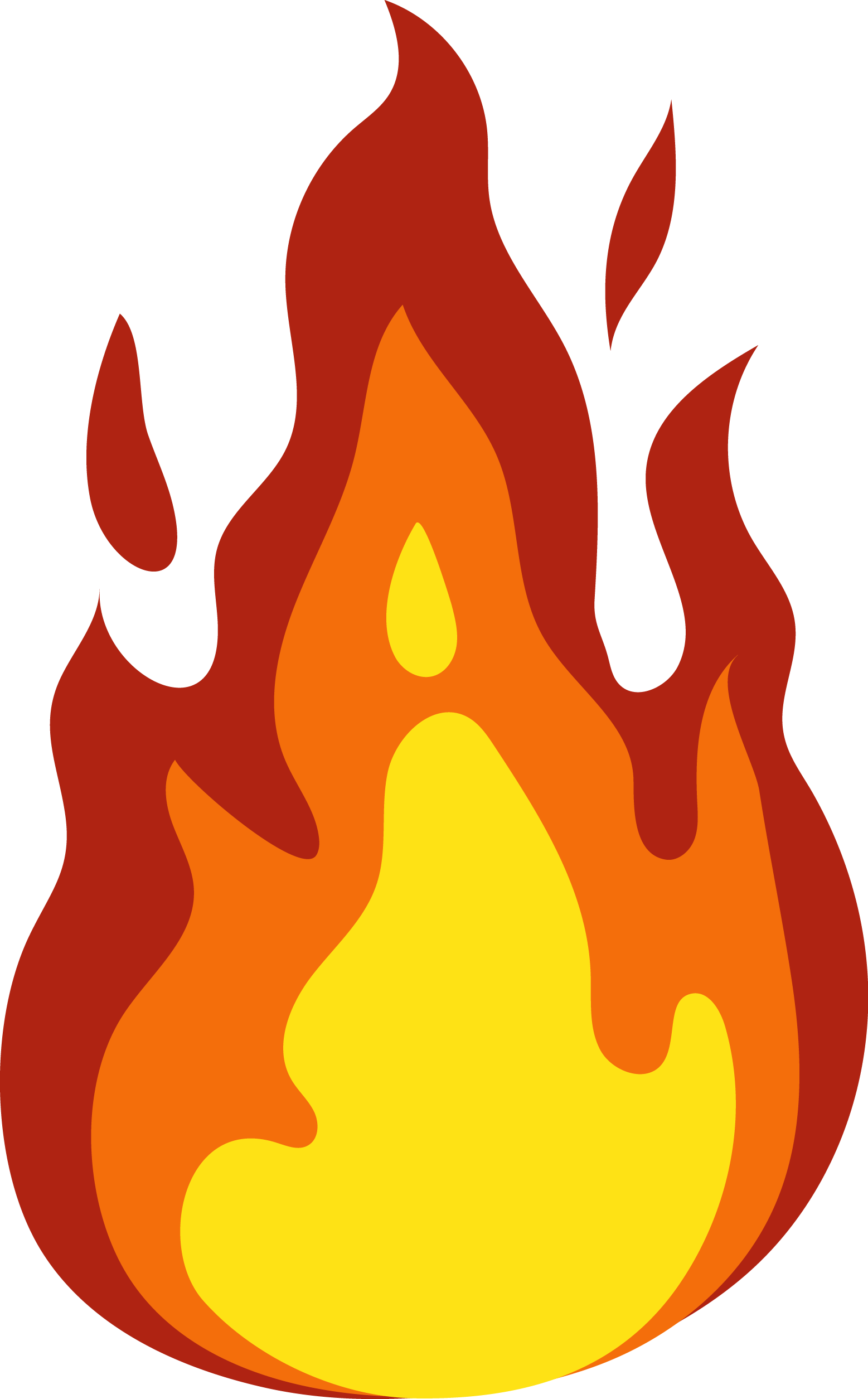} represents the  trainable part.}
	\label{compare}
\end{figure}

Traditional diffusion models typically require powerful GPUs, such as the A100 or H100, and hundreds of GPU days, which constitute the majority of the costs for model training. Inspired by the latent diffusion model \cite{rombach2022high,huang2023composer}, an increasing number of researchers are focusing on the latent space of the image, rather than the pixel space. This approach not only speeds up the processing time but also reduces memory consumption. However, compared to the task of generating images from text, generating videos from text\cite{singer2022make,wu2023tune,wang2023videocomposer,wang2023videofactory,blattmann2023align,blattmann2023stable} requires an additional temporal structure, which incurs further expense. Even when the video diffusion model operates in the latent space, the resource consumption remains considerable.

A conventional T2V framework from text typically consists of three components: Variational Auto-Encoders (VAE), a text encoder, and a 3D denoising Unet. During the training phase, the VAE and text encoder are frozen and initialized with a pre-trained model, leaving only the 3D-Unet to be trained. The 3D-Unet primarily incorporates additional temporal layers compared to the 2D-Unet, enabling its spatial layers’ parameters to be initialized from the well-pre-trained 2D-Unet. Given the available computational resources, training tasks from text to video generation are generally divided into two types. As depicted in Figure \ref{compare}, the first is full fine-tuning, which necessitates updating the parameters of both the original spatial layers and the extra temporal layers. The second is delta-tuning, which involves updating only the parameters of the extra temporal layers. Full fine-tuning requires more GPU memory and training time, but it may compromise the capabilities of the well-trained 2D-Unet model. Delta tuning, on the other hand, may reduce memory usage and training time, but it still retains unnecessary gradients, indicating that there is room for improvement in the existing structure.

To address these challenges, we present Mobius, a novel T2V framework that employs a highly efficient spatial-temporal parallel training paradigm. Despite the significant impact of the gradient computation process on the training process, most existing fine-tuning work \cite{jia2022visual,li2022learning,liu2022prompt,chen2022conv,yin20231,yin2023adapter} does not consider it. Unnecessary and ineffective gradient computations not only increase GPU memory usage but also prolong the training time. Acknowledging this crucial issue, we propose a new 3D-Unet structure. In this structure, \textcolor{red}{\textbf{the gradient flow of the spatial and temporal layers is parallel}}, eliminating the need to maintain the gradient of the frozen spatial layers.

To demonstrate the effectiveness of our new Spatial-Temporal Parallel Training Paradigm, we conducted experiments using our cartoon emojis dataset, which was collected from the Internet. We utilized approximately 330,000 text-GIF pairs for training. The text descriptions were generated by the Video-LLAVA model using cartoon emojis as input. In delta-tuning, our experiments confirmed that the new spatial-temporal parallel framework can save \textcolor{red}{\textbf{24\%}} of GPU memory and \textcolor{red}{\textbf{12\%}} of training time, while achieving video results comparable to the traditional spatial-temporal serial framework. Furthermore, compared to full fine-tuning, our framework offers additional savings in terms of time and memory.

\section{Related Work}
\subsection{Text-to-image Generation}
With the development of the diffusion model, the text-to-image generation task has become a new trend, which can generate satisfactory images. Compared to the basic DDPM\cite{ho2020denoising} and DDIM\cite{song2020denoising}, stable diffusion\cite{rombach2022high} works in the latent space, which can introduce many conditions to generate artificial images with controllable attributes. Meanwhile, it also saves the GPU memory and training time because it adopts efficient and high-dimensional features. Inspired by it, more and more works are proposed, that devise more flexible control modes to generate more accurate images.

ControlNet\cite{zhang2023adding}can add conditions such as edges, depth, segmentation, human pose, etc., to control the image generation of large pre-trained diffusion models. Composer\cite{huang2023composer} is a creative and controllable image synthesis model, which uses a composable condition method, which can flexibly control various attributes of the output image, such as spatial layout and color. Adapter\cite{mou2023t2i} is a novel approach for learning simple and lightweight adapters to align internal knowledge in text-to-image (T2I) models with external control signals (e.g., color and structure). IP-Adapter\cite{ye2023ip} is an effective and lightweight adapter for implementing image cueing for pre-trained text-to-image diffusion models. The key design of the IP adapter is to decouple the cross-attention mechanism, which separates the cross-attention layers for text features and image features.

\subsection{Text-to-video Generation}
Text-to-Video, also known as T2V, is a technology that converts text-based information into video content. It typically employs artificial intelligence algorithms to create videos from provided textual cues.  Compared to text-to-image, text-to-video is more difficult to realize. Recently, some researchers have proposed some technology to solve this issue. Make-A-Video\cite{singer2022make} builds on the T2I model utilizing a novel and efficient space-time module. They decompose the complete temporal U-Net and attention tensor and approximate them in space and time.Tune-A-Video\cite{wu2023tune} proposes a customized spatial-temporal attention mechanism and an efficient one-time adjustment strategy for better learning of continuous motion. For inference, they use DDIM inversion to provide structural guidance for sampling. VideoComposer\cite{wang2023videocomposer} allows users to flexibly compose videos in textual terms, spatial terms, and more importantly, temporal terms. Its key design introduces motion vectors of compressed video as explicit control signals to guide temporal dynamics. VideoFactory\cite{wang2023videofactory} enhances the interaction between spatial and temporal perceptions. Specifically, they use a switched cross-attention mechanism in the 3D window that alternates "querying" roles between spatial and temporal blocks, allowing them to reinforce each other. VideoLDM\cite{blattmann2023align} is applied to high-resolution video generation. They use an LDM pretrained on the image and then transform the image generator into a video generator by introducing the temporal dimension of a latent spatial diffusion model and fine-tuning it on the video.

\section{Methods}
In this section, we will introduce the Mobius framework. Firstly, we will present the preliminaries about the T2V task. Following that, we will demonstrate the innovative design of the spatial-temporal parallel for the 3D-Unet and analyze its high efficiency.

\subsection{Preliminaries}

\begin{figure}[h]
	\centering
	\includegraphics[scale=0.25]{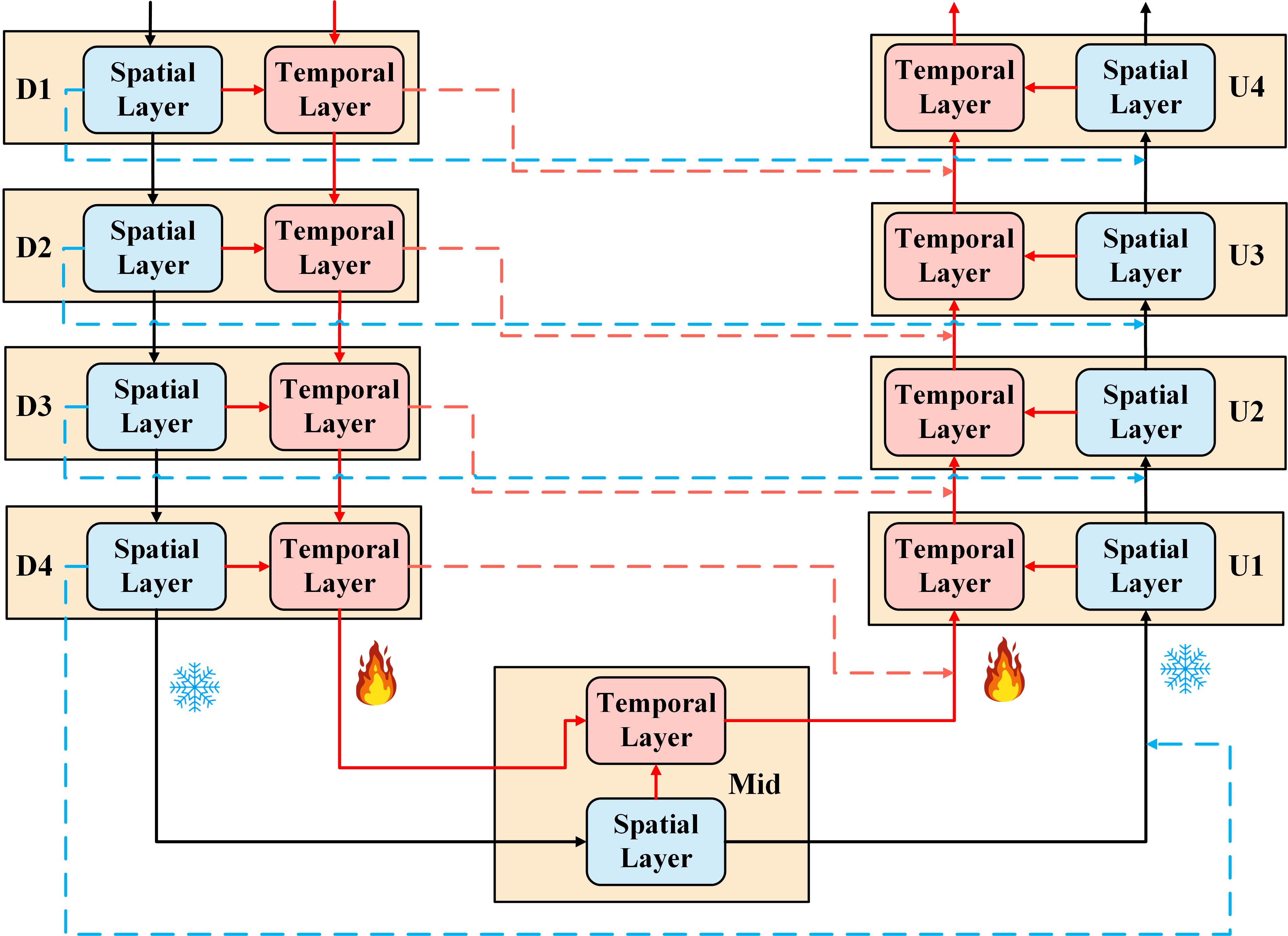}
	\caption{The Mobius framework introduces a spatial-temporal parallel paradigm for text-to-video tasks. Existing text-to-video works often add extra serial temporal layers to ensure continuous action, which can preserve unnecessary gradients. However, the Mobius framework devises a parallel mode for spatial and temporal layers. This means that the gradient only flows in the inner ring, which corresponds to the temporal layers. This approach aims to improve the efficiency and effectiveness of text-to-video generation tasks. During the training process, the spatial layers are frozen, and the temporal layers are trainable.}
	\label{framework}
\end{figure}

\textbf{Diffusion Model}

The diffusion model is a type of latent variable generative model. It primarily consists of three components: a forward process, a reverse process, and a sampling process. The goal of the diffusion model is to learn a diffusion process that generates a probability distribution ($x_0 \sim q(x_0)$ ) for a given dataset. This is achieved by simulating the dispersion of data points in the latent space, thereby learning the underlying structure of the dataset. The model is trained to reverse the process of adding noise to an image. Upon the convergence of training, image generation can be performed by iterative denoising an image composed of random noise. 

Diffusion models are often formulated as Markov chains ($x_1,...,x_T$) and are trained using variational inference ($\beta_t\in(0,1)$). Then the Markov transition $q({x_t}|{x_{t - 1}})$ can be defined as follows, 
\begin{equation}
q({x_t}|{x_{t - 1}}) = N({x_t};\sqrt {1 - {\beta _t}} {x_{t - 1}},{\beta _t}{\mathbb{I}}), t = 1,...,T
\end{equation}
where $\theta$ is the trainable parameters that make sure the generated reverse process is similar to the forward process.

To generate the Markov chain $x_1, . . . , x_T$,  Denoising Diffusion Probabilistic Models utilize the reverse process, which incorporates a prior distribution and Gaussian transitions.
\begin{equation}
{p_\theta }({x_{t - 1}}|{x_t}) = N({x_{t - 1}};{\mu _\theta }({x_t},t)), t = 1,...,T
\end{equation}

Hence, DDPMs maximize the variational lower bound of the negative log-likelihood, which has a closed form, by following the variational inference principle. In reality, we can use a sequence of weight-sharing denoising autoencoders ${\epsilon_\theta }({x_t},t)$, which are well-trained, to predict the denoised variant of their input $x_t$. We give the objective as follows,
\begin{equation}
{E_{x, \epsilon  \sim N(0,1),t}}\left[ {\left\| { \epsilon  - { \epsilon_\theta }({x_t},t)} \right\|_2^2} \right]
\end{equation}

\noindent\textbf{Text-to-video Generation}

Recently, diffusion-based models have been designed to operate in the latent space. This allows for the addition of extra multi-modal conditions and results in a reduction of training costs at the same time.

\begin{figure}[h]
	\centering
	\includegraphics[scale=0.25]{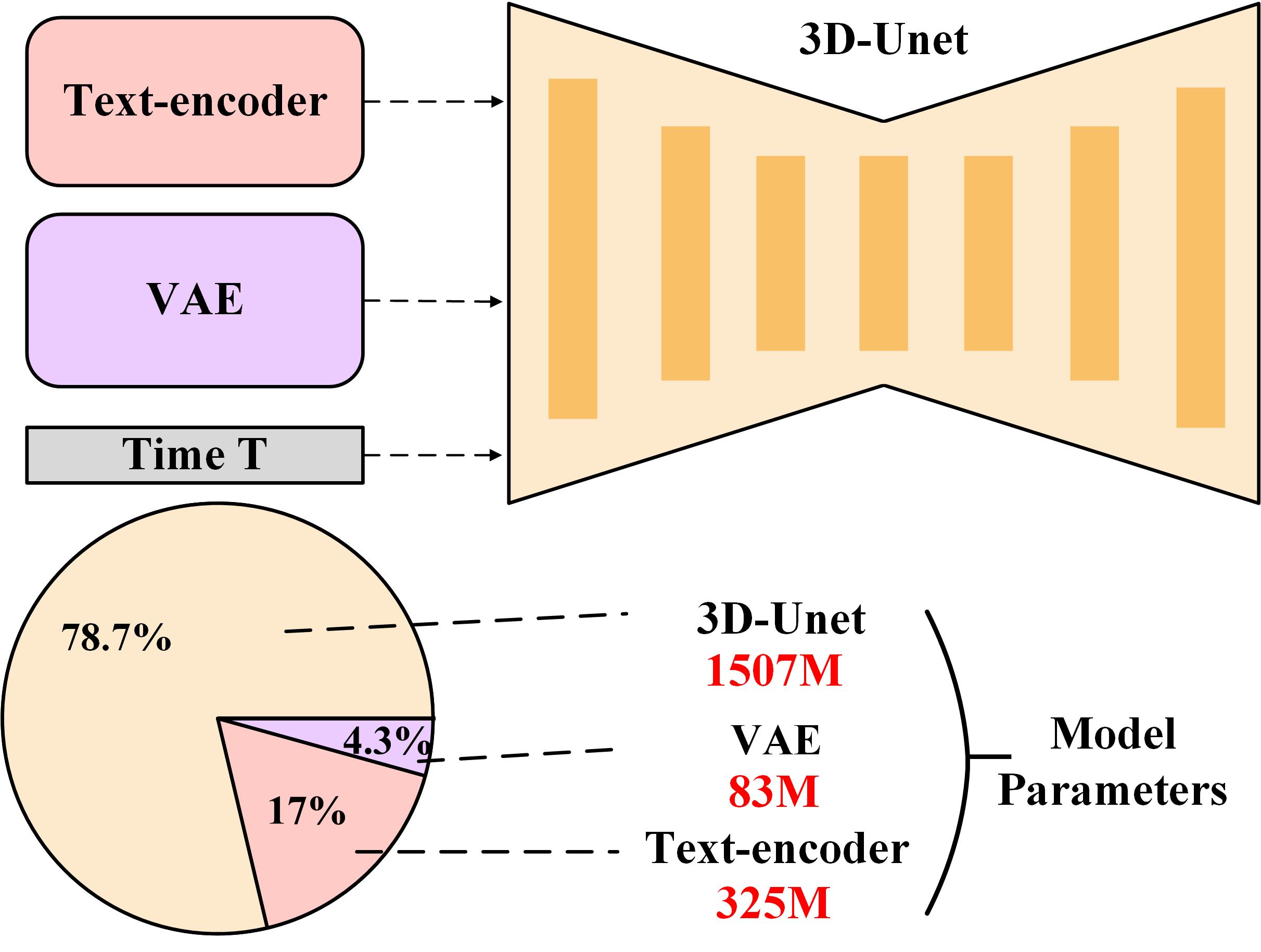}
	\caption{The framework for text-to-video conversion consists of a text-encoder, a VAE, and a 3D-Unet. The pie chart shows that the majority of the model parameters are contained within the 3D-Unet.}
	\label{t2v}
\end{figure}

Most text-to-video models are an extension of text-to-image models, with the addition of temporal layers to ensure the generated images maintain both consistency and motion. As depicted in Fig.\ref{t2v}, the fundamental framework for text-to-video conversion comprises a text-encoder, a Variational Autoencoder (VAE), and a 3D-Unet. The text encoder is tasked with obtaining the prompt encoding, while the VAE is responsible for extracting image features. The 3D-Unet is employed for denoising, representing the network $\epsilon_\theta (x_t, t)$ that was introduced earlier. 

During the training phase, the text-encoder and VAE are frozen, making the 3D-Unet the only trainable component. As demonstrated in the pie chart, the 3D-Unet contains the majority of the model parameters, accounting for a significant \textbf{78.7\%} of the total parameters. Therefore, optimizing the 3D-Unet will significantly improve the training process, including aspects such as GPU memory usage and training time.

\subsection{Mobius}
In this section, we will delve into the details of the Mobius. Given its unique framework, we will present the innovative design in three parts below.

\noindent\textbf{Decouple Hidden State}

In the classic flow, the hidden state $h$ is fed into the 3D-UNet for training the denoising process. The $h$ is obtained by VAE and processed by a pre-network that includes some transformers and convolution modules.

In the devising, we separate the original serial mode into two branches: one for spatial layers and the other for temporal layers. As shown in Fig.\ref{framework}, we provide them with two inputs during the first block. Specifically, the spatial layers continue to use the hidden state $h$. The temporal layers adopt the time state $t$, which is initially copied from $h$. However, in the intermediary network block, it separates from $h$, indicating its decoupling from $h$. This design will make sure the temporal flow will have relative independence, and we will give more analysis in the Section\ref{gradient}.

\noindent\textbf{Double State Bridging}

The previous 3D-Unet employs a bridge structure to fuse the features of the down block and up block for enhanced learning. As depicted in Fig.\ref{framework}, we retain the bridge design for the spatial layers, while also introducing a bridge design for the temporal layers. The double bridge design is shown below,
\begin{equation}
B_{5-i}^U = B^D_i \oplus B^U_{i-2}, i=1,2,3, B\in\{T, U\} 
\end{equation}
\begin{equation}
B_{1}^U = B^D_4 \oplus B^M_{0}, B\in\{T, U\} 
\end{equation}
where $U$, $D$, $M$ represent the up-block, down-block, and middle-block in 3D-Unet, separately. And the $\oplus$ means the concatenate operation. $T$ and $U$ represent the temporal and spatial layers.

\begin{table}[t]
	\centering
	\caption{Definition}
	\setlength{\tabcolsep}{6mm}{
		\begin{tabular}{c|c}
			\toprule
			Symbol & Definition  \\
			\midrule
			$SC$  & Spatial Convolution Block \\
			$TC$  & Temporal Convolution Block \\
			$SA$  & Spatial Attention Block\\
			$TA$  & Temporal Attention Block\\
			$h$  & Hidden State Feature \\
			$t$  & Time State Feature  \\
			\bottomrule
	\end{tabular}}%
	\label{def}%
\end{table}%

\noindent\textbf{Spatial-temporal State Fusion}

For the final output, we employ a lightweight convolution network to merge the spatial and temporal states. The fusion is presented below,
\begin{equation}
f = Conv(h\oplus t)
\end{equation}

This ($f$) is then fed into a post-processing network to obtain the denoising results for the subsequent DDIM operation. 

\begin{figure}[h]
	\centering
	\includegraphics[scale=0.18]{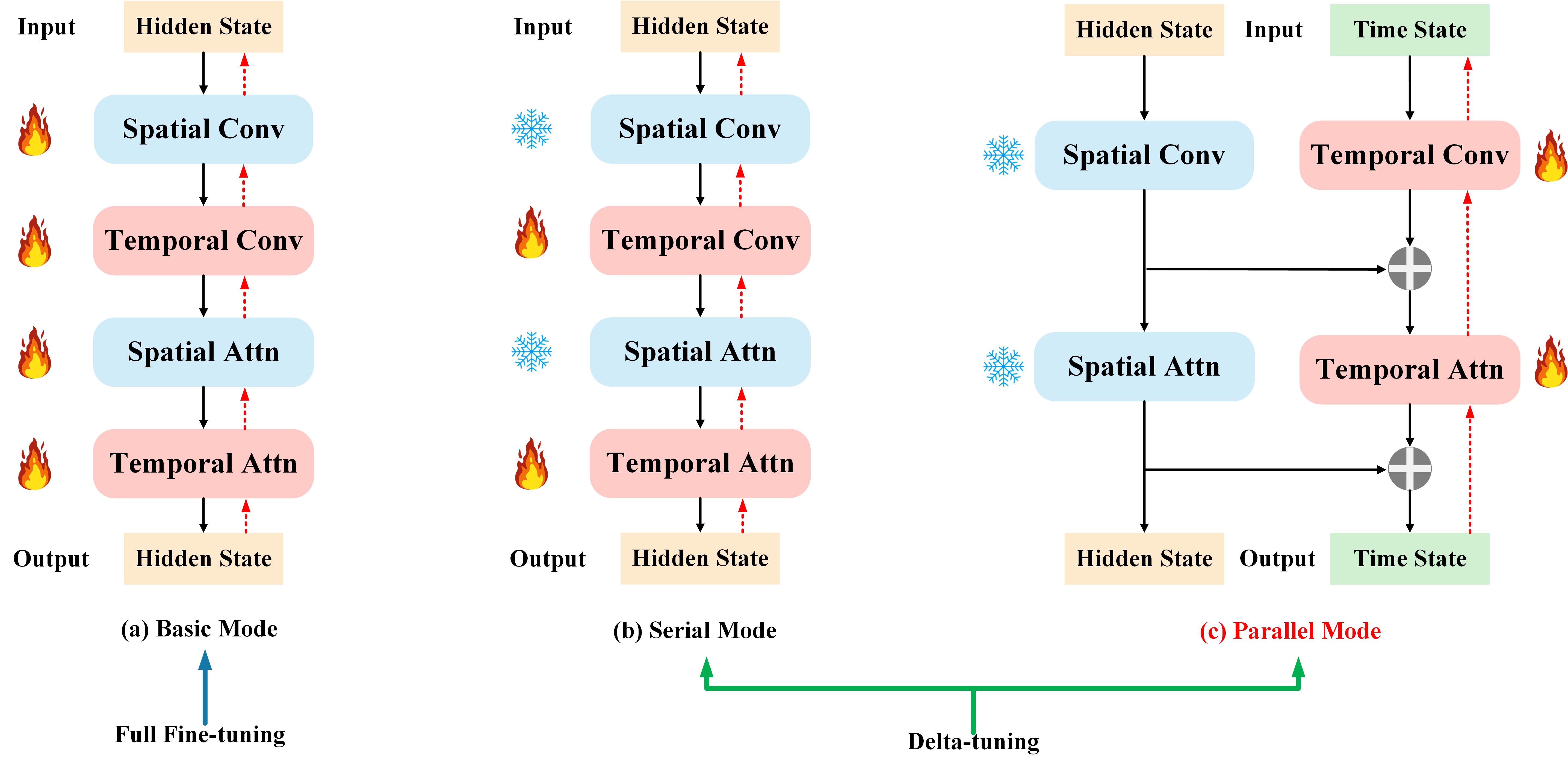}
	\caption{Existing tuning paradigm in text-generate video task.  (a) is the Full fine-tuning mode. (b) and (c) are Delta-tuning. (b) is the serial mode mode, and (c) is our parallel mode.  The black arrow means the forward step, and the red arrow means the backward step. Both modes (a) and (b) need to reserve and compute all the gradients. However, our parallel mode only needs to reserve and compute the temporal layers gradient, which is highly efficient. \protect\includegraphics[height=0.3cm]{./fig/ice.png} represents the frozen part, and \protect\includegraphics[height=0.3cm]{./fig/fire.png} represents the  trainable part.}
	\label{training}
\end{figure}

\subsection{Comparison with Serial Mode}
\label{gradient}
In this part, we will discuss the differences between serial and parallel modes by analyzing the forward propagation process. We will also explain how the parallel mode can significantly reduce the GPU memory usage during the gradient backward propagation process.

\noindent\textbf{Forward Propagation}

As illustrated in Fig. \ref{training}, we demonstrate the forward propagation of three tuning modes. For a more comprehensive analysis, we initially provide some definitions in Table \ref{def}.

For the $i$-th block, the output $h_{i+1}$ of the serial mode fine-tuning can be written as follows:
\begin{equation}
h_{i+1} = TA(SA(TC(SC(h_i))))
\label{bi}
\end{equation}

Our designed parallel mode has pair of (input, output), ($h_i$, $t_i$), and  the forward process of $h$ and $t$ can be represented as follows:
\begin{equation}
h_{i+1} = SA(SC(h_i))
\end{equation}
\begin{equation}
t_{i+1} = SA(SC(h_i)) + TA(SC(h_i)+TC(t_i))
\label{ti}
\end{equation}

It is clear that Eq.\ref{bi} and Eq.\ref{ti} have different flows in the gradient. The serial mode inserts the temporal layers into the basic net, so the subsequent calculation depends on the preceding spatial and temporal layers. Thus, the update of the temporal layers depends on the backward propagation through the entire network. In contrast, the update in parallel mode does not need to depend on the backward propagation of the entire network.

Next, we will analyze both processes for comparison. To simplify the equation, we will introduce some conversions.  During the training process, $h_{i}$ and $t_{i}$ have the same feature shape, so we define them as $w_i$. Moreover, $SA$ and $TA$ have similar parameters, so we define its operation as $\alpha$. $SC$ and $TC$ also have similar parameters, so we define its operation as $\beta$.

So we can rewrite the Eq.\ref{ti} as follows,
\begin{equation}
	\begin{aligned}
		w_{i+1} & = SA(SC(h_i)) +  TA(SC(h_i) + TA(TC(t_i)) \\
            & = 3\alpha(\beta(w_i))
	\end{aligned}
\label{ti_new}
\end{equation}

And the Eq.\ref{bi} can be transferred as below,
\begin{equation}
w_{i+1}=\alpha(\beta(\alpha(\beta(w_i))))
\label{bi_new}
\end{equation}

When considering the computation of $\alpha$ and $\beta$, Eq.\ref{ti_new} involves fewer and simpler operations compared to Eq.\ref{bi_new}, which demonstrates the efficient of the parallel mode.

\noindent\textbf{Backward Propagation}

We will derive the process for calculating the gradient of the parameters $\theta_i=\{\vartheta_i^1, \vartheta_i^2, \vartheta_i^3,...,\vartheta_i^n\}$ for the temporal layer in the $i$-th block. In this context, $L$ represents the loss, and $f_S$ denotes the function of the temporal layer in the $i$-th block. Additionally, the output of the $i$-th layer is represented as $p_{i+1}$.

Therefore, we give the partial derivative of the $L$ with respect to $\theta_i$, 
\begin{equation}
	\begin{aligned}
\frac{{\partial L}}{{\partial {\theta _i}}} = \frac{{\partial L}}{{\partial {p_{i + 1}}}}\frac{{\partial {p_{i + 1}}}}{{\partial {f_S}}}\frac{{\partial {f_S}}}{{\partial {\theta _i}}}
	\end{aligned}
\label{back_new}
\end{equation}

To better analyze the Eq.\ref{back_new}, we give the partial derivative of $p_{i+1}$ with respect to $f_S$,
\begin{equation}
	\begin{aligned}
		\frac{{\partial {p_{i + 1}}}}{{\partial {f_S}}} = \frac{{\partial {p_{i + 1}}}}{{\partial f_i^1}}\frac{{\partial f_i^1}}{{\partial f_i^2}}\frac{{\partial f_i^2}}{{\partial f_i^3}} \cdots \frac{{\partial f_i^m}}{{\partial {f_S}}}
	\end{aligned}
\label{back}
\end{equation}

Hence the Eq.\ref{back_new} can be rewritten as follows,
\begin{equation}
	\begin{aligned}
\frac{{\partial L}}{{\partial {\theta _i}}} = \frac{{\partial L}}{{\partial {p_{i + 1}}}}\left( {\frac{{\partial {p_{i + 1}}}}{{\partial f_i^1}}\frac{{\partial f_i^1}}{{\partial f_i^2}}\frac{{\partial f_i^2}}{{\partial f_i^3}} \cdots \frac{{\partial f_i^m}}{{\partial {f_S}}}} \right)\sum\limits_{k = 1}^n {\frac{{\partial {f_S}}}{{\partial \vartheta _i^k}}} 
	\end{aligned}
\label{back_new2}
\end{equation}

The serial mode only reduces the trainable model parameters, but it does not reduce the intermediate process $\frac{{\partial {p_{i + 1}}}}{{\partial {f_S}}}$. Our parallel mode significantly reduces $\frac{{\partial {p_{i + 1}}}}{{\partial {f_S}}}$ with efficient gradient flow, which can save GPU memory and training time.

\section{Experiments}

\begin{figure}[t]
	\centering
	\includegraphics[scale=0.15]{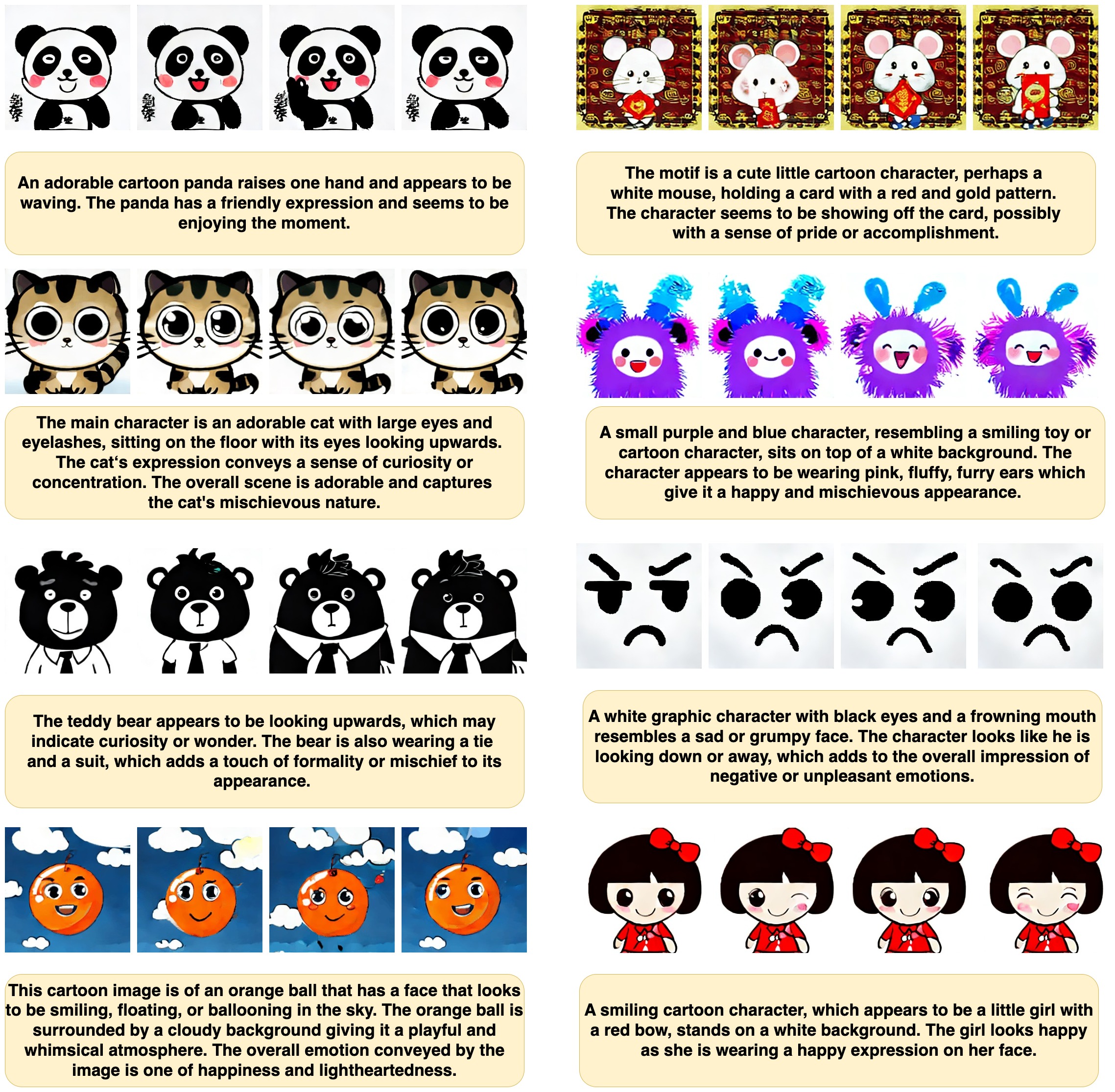}
	\caption{Prompt words (Chinese are translated to English for reading convenience.) and their corresponding generated video}
	\label{demo}
\end{figure}

\subsection{Implementation Details}

\textbf{Dataset}:

We have collected 6.8 million emoji images submitted to the Emoji Store, including 0.62 million static images and 6.18 million GIF motion pictures, each with a text tag that expresses a social emotion. Then we use the LLaVa\cite{liu2023visual} to obtain the image caption of each image (or the first frame of the GIF motion pictures). And Video-LLaMA\cite{zhang2023video} is adopted to acquire the video caption of the motion image.  However, the video captions were too long and prone to be truncated by the text encoder, so we use Llama2\cite{touvron2023llama} to do the summarization extraction. Finally, we translated both image captions and summarized video captions into Chinese, which is suitable for business scenarios. Since most of the images contain Chinese text, which will reduce the quality of generation and make the result with garbled code, we use OCR to detect the text area and then use LaMa\cite{suvorov2022resolution} to paint the text into the image. After this series of processing, we obtained 6.8 million text-removed image sets with social sentiment tags, image captions, and video captions.

\noindent\textbf{Training Details}:

\textbf{Stage 1: Text-to-image Generation}

The VAE is initialized from the original Stable Diffusion 2.1\footnote{https://huggingface.co/stabilityai/stable-diffusion-2-1}. And the text encoder adopts the Chinese CLIP text encoder chinese-clip-vit-huge-patch14\footnote{https://github.com/OFA-Sys/Chinese-CLIP}. And the Unet is trained from scratch based on the emoji dataset that we collected above. We use the first frame of all the static and motion images as the training set, and the image caption is adopted as the text label.

\textbf{Stage 2: Text-to-video Generation}

In the above emoji packet dataset, we retained the motion pictures with a frame count greater than 4 frames, and removed the pictures with a larger text area, leaving about 1 million of motion picture data;  we observed that the action patterns in the motion pictures with the same emotion labels tend to be similar, so to facilitate the model to learn the high-quality and smooth actions, as well as to facilitate the quick validation of the generation effect of the motion pictures, we only retained the pictures corresponding to the high-frequency (frequency>1000) social emotion labels, and finally obtained 330 thousand of motion pictures for training.

\subsection{Main Results}
In this part, we will present some visualization generated by our model. We will show the generated effect in two forms, one is to extract the keyframes from the generated video and show them, and the other is to show its dynamic effect directly. For the best visualization, \textbf{we strongly recommend two PDF readers}, which can show the dynamic gif in PDF. One is the  \textcolor{red}{Adobe Acrobat/ Reader\footnote{https://get.adobe.com/reader/}}, and the other is \textcolor{red}{Foxit Reader\footnote{https://www.foxitsoftware.cn/pdf-reader/?agent=foxit\&MD=menu}.} 


\textbf{Static Sequential Frame Display:}As shown in Fig.\ref{demo}, we present the prompt words and their corresponding generated motion picture. Our model will infer 8 frames for each video and we only show 4 key frames to display its dynamic results. Owing to our model being supported for Chinese, we translate the Chinese to English prompts in Fig.\ref{demo} for reading convenience. As shown in Fig.\ref{demo}, the generated video has a nice main body and coherence for the motion. No matter the eyes blink or hand up and down movement all present reasonable results. 


\textbf{Dynamic Sequential Frame Display:}In this section, we display the GIFs that are only supported by the above two PDF readers. As shown below, we compare the video generation results of serial mode and parallel mode. The parallel mode has competitive results compared to the serial mode, which shows its potential. We will provide more efficiency analysis to confirm the advantage in the next section.

\textcolor{red}{\textbf{Serial Mode}}

\begin{center}
  \animategraphics[width=2.5cm,height=2.5cm, autoplay, loop]{10}{./fig_serial/a-}{1}{8}\animategraphics[width=2.5cm,height=2.5cm, autoplay, loop]{10}{./fig_serial/b-}{1}{8}\animategraphics[width=2.5cm,height=2.5cm, autoplay, loop]{10}{./fig_serial/c-}{1}{8}\animategraphics[width=2.5cm,height=2.5cm, autoplay, loop]{10}{./fig_serial/1-}{1}{8}
\end{center}

\begin{center}
  \animategraphics[width=2.5cm,height=2.5cm, autoplay, loop]{10}{./fig_serial/d-}{1}{8}\animategraphics[width=2.5cm,height=2.5cm, autoplay, loop]{10}{./fig_serial/e-}{1}{8}\animategraphics[width=2.5cm,height=2.5cm, autoplay, loop]{10}{./fig_serial/f-}{1}{8}\animategraphics[width=2.5cm,height=2.5cm, autoplay, loop]{10}{./fig_serial/2-}{1}{8}
\end{center}

\textcolor{red}{\textbf{Parallel Mode}}

\begin{center}
  \animategraphics[width=2.5cm,height=2.5cm, autoplay, loop]{10}{./fig2/16-}{1}{8}\animategraphics[width=2.5cm,height=2.5cm, autoplay, loop]{10}{./fig2/8-}{1}{8}\animategraphics[width=2.5cm,height=2.5cm, autoplay, loop]{10}{./fig2/17-}{1}{8}\animategraphics[width=2.5cm,height=2.5cm, autoplay, loop]{10}{./fig2/27-}{1}{8}
\end{center}

\begin{center}
  \animategraphics[width=2.5cm,height=2.5cm, autoplay, loop]{10}{./fig2/18-}{1}{8}\animategraphics[width=2.5cm,height=2.5cm, autoplay, loop]{10}{./fig2/19-}{1}{8}\animategraphics[width=2.5cm,height=2.5cm, autoplay, loop]{10}{./fig2/20-}{1}{8}\animategraphics[width=2.5cm,height=2.5cm, autoplay, loop]{10}{./fig2/5-}{1}{5}
\end{center}

\begin{center}
  \animategraphics[width=2.5cm,height=2.5cm, autoplay, loop]{10}{./fig2/21-}{1}{8}\animategraphics[width=2.5cm,height=2.5cm, autoplay, loop]{10}{./fig2/22-}{1}{8}\animategraphics[width=2.5cm,height=2.5cm, autoplay, loop]{10}{./fig2/23-}{1}{8} \animategraphics[width=2.5cm,height=2.5cm, autoplay, loop]{10}{./fig2/29-}{1}{8}
\end{center}

\begin{center}
  \animategraphics[width=2.5cm,height=2.5cm, autoplay, loop]{10}{./fig2/6-}{1}{8}\animategraphics[width=2.5cm,height=2.5cm, autoplay, loop]{10}{./fig2/25-}{1}{8}\animategraphics[width=2.5cm,height=2.5cm, autoplay, loop]{10}{./fig2/26-}{1}{8}\animategraphics[width=2.5cm,height=2.5cm, autoplay, loop]{10}{./fig2/1-}{1}{8}
\end{center}

\begin{center}
   \animategraphics[width=2.5cm,height=2.5cm, autoplay, loop]{10}{./fig2/2-}{1}{8}\animategraphics[width=2.5cm,height=2.5cm, autoplay, loop]{10}{./fig2/3-}{1}{8}\animategraphics[width=2.5cm,height=2.5cm, autoplay, loop]{10}{./fig2/4-}{1}{8}\animategraphics[width=2.5cm,height=2.5cm, autoplay, loop]{10}{./fig2/7-}{1}{8}
\end{center}

\begin{center}
   \animategraphics[width=2.5cm,height=2.5cm, autoplay, loop]{10}{./fig2/24-}{1}{8}\animategraphics[width=2.5cm,height=2.5cm, autoplay, loop]{10}{./fig2/27-}{1}{8}\animategraphics[width=2.5cm,height=2.5cm, autoplay, loop]{10}{./fig2/28-}{1}{8}\animategraphics[width=2.5cm,height=2.5cm, autoplay, loop]{10}{./fig2/13-}{1}{8}
\end{center}

\begin{center}
   \animategraphics[width=2.5cm,height=2.5cm, autoplay, loop]{10}{./fig2/12-}{1}{8}\animategraphics[width=2.5cm,height=2.5cm, autoplay, loop]{10}{./fig2/14-}{1}{8}\animategraphics[width=2.5cm,height=2.5cm, autoplay, loop]{10}{./fig2/15-}{1}{8}\animategraphics[width=2.5cm,height=2.5cm, autoplay, loop]{10}{./fig2/11-}{1}{8}
\end{center}

\begin{center}
   \animategraphics[width=2.5cm,height=2.5cm, autoplay, loop]{10}{./fig2/10-}{1}{8}\animategraphics[width=2.5cm,height=2.5cm, autoplay, loop]{10}{./fig2/30-}{1}{8}\animategraphics[width=2.5cm,height=2.5cm, autoplay, loop]{10}{./fig2/31-}{1}{8}\animategraphics[width=2.5cm,height=2.5cm, autoplay, loop]{10}{./fig2/32-}{1}{8}
\end{center}

\begin{center}
   \animategraphics[width=2.5cm,height=2.5cm, autoplay, loop]{10}{./fig2/33-}{1}{8}\animategraphics[width=2.5cm,height=2.5cm, autoplay, loop]{10}{./fig2/34-}{1}{8}\animategraphics[width=2.5cm,height=2.5cm, autoplay, loop]{10}{./fig2/35-}{1}{8}\animategraphics[width=2.5cm,height=2.5cm, autoplay, loop]{10}{./fig2/36-}{1}{8}
\end{center}

\subsection{Efficiency Analysis}
In this part, we will demonstrate the high efficiency of our parallel mode compared to the traditional serial mode. During the experiment, we use 4 Nvidia A100 GPUs with 40GB of memory each. We train the model for 1 million optimization steps. For each GPU, we set the batch size to 2, and the gradient accumulation step to 2. Thus, the total batch size is 16.

\begin{figure}[h]
	\centering
	\includegraphics[scale=0.43]{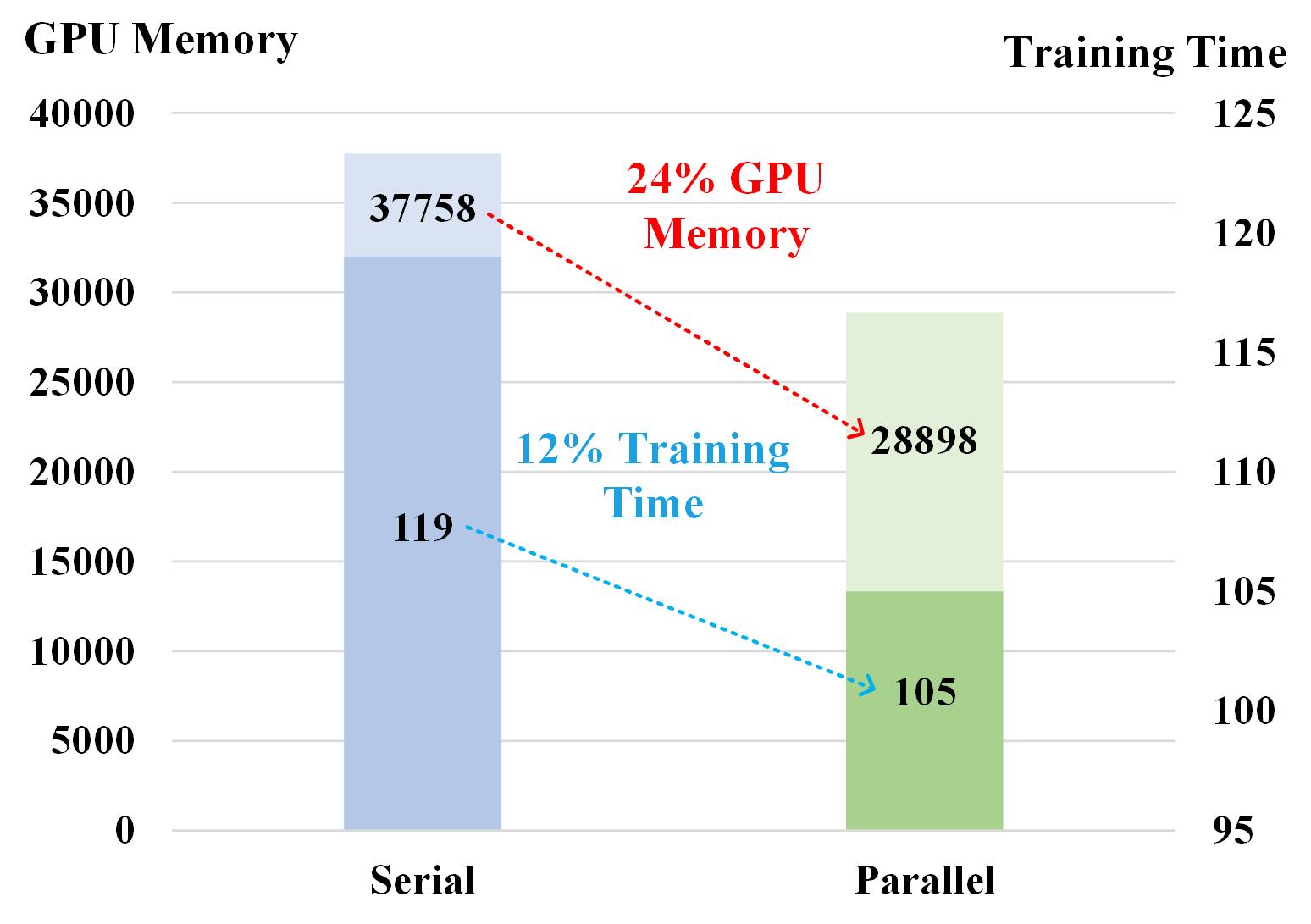}
	\caption{Comparisons of GPU memory and training time between serial mode and parallel mode. We only show 10w steps for training time for a clear presentation.}
	\label{effi}
\end{figure}

As shown in Fig.\ref{effi}, we compare the GPU memory usage and training time between the two modes. The classical serial mode requires 37758 MB memory, which means it can only be trained on A100 with 40960 MB at least. However, our proposed parallel mode only needs 28898 MB memory, which means we can train it on cheaper V100 with 32768 MB. Our spatial-temporal parallel mode can save \textcolor{red}{\textbf{24\%}} GPU memory compared to the baseline. Moreover, we also report the training time of 100000 training steps. The serial mode takes 119 hours, but our parallel mode takes 105 hours, which means our method can save \textcolor{red}{\textbf{12\%}} of time.

\subsection{Limitation}
Although our spatial-temporal parallel mode deviates from the traditional flow, there is still room for some new designs to enhance the spatial-temporal feature fusion, which will produce more coherent videos.

\section{Conclusions}
In this paper, we propose a highly efficient spatial-temporal parallel training paradigm for text-to-video generation tasks, named Mobius. The traditional text-to-video model usually starts from a well-trained text-to-image model, and it only needs to fine-tune the extra temporal layers in the new 3D-Unet. However, the previous 3D-Unet is a serial mode, which means the temporal layers follow the spatial layers. This design makes the model retain unnecessary parameters and gradients, which leads to high training time and GPU memory consumption. Our parallel mode effectively solves this problem and saves \textcolor{red}{\textbf{24\%}} GPU memory and \textcolor{red}{\textbf{12\%}} training time. We hope our work will offer some insights for the large model fine-tuning in the AIGC community.


\clearpage  

%
%
\bibliographystyle{splncs04}
\bibliography{main}

\begin{thebibliography}{10}
\providecommand{\url}[1]{\texttt{#1}}
\providecommand{\urlprefix}{URL }
\providecommand{\doi}[1]{https://doi.org/#1}

\bibitem{blattmann2023stable}
Blattmann, A., Dockhorn, T., Kulal, S., Mendelevitch, D., Kilian, M., Lorenz, D., Levi, Y., English, Z., Voleti, V., Letts, A., et~al.: Stable video diffusion: Scaling latent video diffusion models to large datasets. arXiv preprint arXiv:2311.15127  (2023)

\bibitem{blattmann2023align}
Blattmann, A., Rombach, R., Ling, H., Dockhorn, T., Kim, S.W., Fidler, S., Kreis, K.: Align your latents: High-resolution video synthesis with latent diffusion models. In: Proceedings of the IEEE/CVF Conference on Computer Vision and Pattern Recognition. pp. 22563--22575 (2023)

\bibitem{brock2018large}
Brock, A., Donahue, J., Simonyan, K.: Large scale gan training for high fidelity natural image synthesis. arXiv preprint arXiv:1809.11096  (2018)

\bibitem{chen2022conv}
Chen, H., Tao, R., Zhang, H., Wang, Y., Ye, W., Wang, J., Hu, G., Savvides, M.: Conv-adapter: Exploring parameter efficient transfer learning for convnets. arXiv preprint arXiv:2208.07463  (2022)

\bibitem{chen2020generative}
Chen, M., Radford, A., Child, R., Wu, J., Jun, H., Luan, D., Sutskever, I.: Generative pretraining from pixels. In: International conference on machine learning. pp. 1691--1703. PMLR (2020)

\bibitem{child2019generating}
Child, R., Gray, S., Radford, A., Sutskever, I.: Generating long sequences with sparse transformers. arXiv preprint arXiv:1904.10509  (2019)

\bibitem{goodfellow2020generative}
Goodfellow, I., Pouget-Abadie, J., Mirza, M., Xu, B., Warde-Farley, D., Ozair, S., Courville, A., Bengio, Y.: Generative adversarial networks. Communications of the ACM  \textbf{63}(11),  139--144 (2020)

\bibitem{ho2020denoising}
Ho, J., Jain, A., Abbeel, P.: Denoising diffusion probabilistic models. Advances in neural information processing systems  \textbf{33},  6840--6851 (2020)

\bibitem{huang2023composer}
Huang, L., Chen, D., Liu, Y., Shen, Y., Zhao, D., Zhou, J.: Composer: Creative and controllable image synthesis with composable conditions. arXiv preprint arXiv:2302.09778  (2023)

\bibitem{jia2022visual}
Jia, M., Tang, L., Chen, B.C., Cardie, C., Belongie, S., Hariharan, B., Lim, S.N.: Visual prompt tuning. In: European Conference on Computer Vision. pp. 709--727. Springer (2022)

\bibitem{li2022learning}
Li, A., Zhuang, L., Fan, S., Wang, S.: Learning common and specific visual prompts for domain generalization. In: Proceedings of the Asian Conference on Computer Vision. pp. 4260--4275 (2022)

\bibitem{liu2023visual}
Liu, H., Li, C., Wu, Q., Lee, Y.J.: Visual instruction tuning. arXiv preprint arXiv:2304.08485  (2023)

\bibitem{liu2022prompt}
Liu, L., Chang, J., Yu, B.X., Lin, L., Tian, Q., Chen, C.W.: Prompt-matched semantic segmentation. arXiv preprint arXiv:2208.10159  (2022)

\bibitem{mou2023t2i}
Mou, C., Wang, X., Xie, L., Zhang, J., Qi, Z., Shan, Y., Qie, X.: T2i-adapter: Learning adapters to dig out more controllable ability for text-to-image diffusion models. arXiv preprint arXiv:2302.08453  (2023)

\bibitem{van2016conditional}
Van~den Oord, A., Kalchbrenner, N., Espeholt, L., Vinyals, O., Graves, A., et~al.: Conditional image generation with pixelcnn decoders. Advances in neural information processing systems  \textbf{29} (2016)

\bibitem{ramesh2021zero}
Ramesh, A., Pavlov, M., Goh, G., Gray, S., Voss, C., Radford, A., Chen, M., Sutskever, I.: Zero-shot text-to-image generation. In: International Conference on Machine Learning. pp. 8821--8831. PMLR (2021)

\bibitem{razavi2019generating}
Razavi, A., Van~den Oord, A., Vinyals, O.: Generating diverse high-fidelity images with vq-vae-2. Advances in neural information processing systems  \textbf{32} (2019)

\bibitem{rombach2022high}
Rombach, R., Blattmann, A., Lorenz, D., Esser, P., Ommer, B.: High-resolution image synthesis with latent diffusion models. In: Proceedings of the IEEE/CVF conference on computer vision and pattern recognition. pp. 10684--10695 (2022)

\bibitem{singer2022make}
Singer, U., Polyak, A., Hayes, T., Yin, X., An, J., Zhang, S., Hu, Q., Yang, H., Ashual, O., Gafni, O., et~al.: Make-a-video: Text-to-video generation without text-video data. arXiv preprint arXiv:2209.14792  (2022)

\bibitem{song2020denoising}
Song, J., Meng, C., Ermon, S.: Denoising diffusion implicit models. arXiv preprint arXiv:2010.02502  (2020)

\bibitem{suvorov2022resolution}
Suvorov, R., Logacheva, E., Mashikhin, A., Remizova, A., Ashukha, A., Silvestrov, A., Kong, N., Goka, H., Park, K., Lempitsky, V.: Resolution-robust large mask inpainting with fourier convolutions. In: Proceedings of the IEEE/CVF winter conference on applications of computer vision. pp. 2149--2159 (2022)

\bibitem{touvron2023llama}
Touvron, H., Martin, L., Stone, K., Albert, P., Almahairi, A., Babaei, Y., Bashlykov, N., Batra, S., Bhargava, P., Bhosale, S., et~al.: Llama 2: Open foundation and fine-tuned chat models. arXiv preprint arXiv:2307.09288  (2023)

\bibitem{van2016pixel}
Van Den~Oord, A., Kalchbrenner, N., Kavukcuoglu, K.: Pixel recurrent neural networks. In: International conference on machine learning. pp. 1747--1756. PMLR (2016)

\bibitem{wang2023videofactory}
Wang, W., Yang, H., Tuo, Z., He, H., Zhu, J., Fu, J., Liu, J.: Videofactory: Swap attention in spatiotemporal diffusions for text-to-video generation. arXiv preprint arXiv:2305.10874  (2023)

\bibitem{wang2023videocomposer}
Wang, X., Yuan, H., Zhang, S., Chen, D., Wang, J., Zhang, Y., Shen, Y., Zhao, D., Zhou, J.: Videocomposer: Compositional video synthesis with motion controllability. arXiv preprint arXiv:2306.02018  (2023)

\bibitem{wu2023tune}
Wu, J.Z., Ge, Y., Wang, X., Lei, S.W., Gu, Y., Shi, Y., Hsu, W., Shan, Y., Qie, X., Shou, M.Z.: Tune-a-video: One-shot tuning of image diffusion models for text-to-video generation. In: Proceedings of the IEEE/CVF International Conference on Computer Vision. pp. 7623--7633 (2023)

\bibitem{ye2023ip}
Ye, H., Zhang, J., Liu, S., Han, X., Yang, W.: Ip-adapter: Text compatible image prompt adapter for text-to-image diffusion models. arXiv preprint arXiv:2308.06721  (2023)

\bibitem{yin2023adapter}
Yin, D., Hu, L., Li, B., Zhang, Y.: Adapter is all you need for tuning visual tasks. arXiv e-prints pp. arXiv--2311 (2023)

\bibitem{yin20231}
Yin, D., Yang, Y., Wang, Z., Yu, H., Wei, K., Sun, X.: 1\% vs 100\%: Parameter-efficient low rank adapter for dense predictions. In: Proceedings of the IEEE/CVF Conference on Computer Vision and Pattern Recognition. pp. 20116--20126 (2023)

\bibitem{zhang2023video}
Zhang, H., Li, X., Bing, L.: Video-llama: An instruction-tuned audio-visual language model for video understanding. arXiv preprint arXiv:2306.02858  (2023)

\bibitem{zhang2023adding}
Zhang, L., Rao, A., Agrawala, M.: Adding conditional control to text-to-image diffusion models. In: Proceedings of the IEEE/CVF International Conference on Computer Vision. pp. 3836--3847 (2023)

\end{thebibliography}
\end{document}